\documentclass[sigconf]{acmart}
\usepackage{booktabs, multirow, makecell,graphicx,float,balance}

\usepackage{amssymb}

\graphicspath{{./}{study/}}

\AtBeginDocument{%
  }

\copyrightyear{2026}
\acmYear{2026}
\setcopyright{cc}
\setcctype{by-nc-nd}
\acmConference[MM '26]
{Proceedings of the 34th ACM International Conference on Multimedia}
{November 10--14, 2026}
{Rio de Janeiro, Brazil.}
\acmBooktitle{Proceedings of the 34th ACM International Conference on Multimedia
	(MM '26), November 10--14, 2026, Rio de Janeiro, Brazil}
\acmISBN{979-8-4007-2213-4/2026/11}
\acmDOI{10.1145/3767308.3835650}
\begin{document}

\title{EmoZone-Talker: Regional Semantic Control of Audio-Driven 3DGS Talking Heads via Facial Action Units}

\author{Tingting Chen}
\authornote{Equal contribution.}
\affiliation{%
	\institution{China University of Petroleum (East China)}
	\city{Qingdao}
	\country{China}
}
\email{2217010215@s.upc.edu.cn}

\author{Shaojun Wang}
\authornotemark[1]
\affiliation{%
	\institution{China University of Petroleum (East China)}
	\city{Qingdao}
	\country{China}
}
\email{2307010404@s.upc.edu.cn}

\author{Huaye Zhang}
\authornotemark[1]
\affiliation{%
	\institution{China University of Petroleum (East China)}
	\city{Qingdao}
	\country{China}
}
\email{2306030225@s.upc.edu.cn}

\author{Diqiong Jiang}
\authornote{Corresponding author.}
\affiliation{%
	\institution{China University of Petroleum (East China)}
	\city{Qingdao}
	\country{China}
}
\email{djiang@upc.edu.cn}

\author{Chenglizhao Chen}
\affiliation{%
	\institution{China University of Petroleum (East China)}
	\city{Qingdao}
	\country{China}
}
\email{cclz123@163.com}

\renewcommand{\shortauthors}{Tingting Chen, Shaojun Wang, Huaye Zhang, Diqiong Jiang, and Chenglizhao Chen}

\begin{abstract}
3D Gaussian Splatting (3DGS) supports high-fidelity talking head synthesis, but fine-grained regional expression control remains difficult when speech-driven motion and explicit Action Unit (AU) controls affect overlapping facial regions. Frame-level AU trajectories may also contain high-frequency noise that produces unstable deformation. We present EmoZone-Talker, an AU-conditioned framework that treats these issues as spatial responsibility allocation and temporal conditioning. Synergy Zones with Prioritized Attention Bias (SZ-PAB) introduces soft region-aware priors that bias audio toward articulation regions and AUs toward upper-face regions while retaining a learnable transition zone. The Channel-Independent Temporal AU Encoder (CIT-AE) applies learnable local temporal filtering before multimodal fusion. Integrated with 3D Gaussian deformation, these components improve measured AU controllability and temporal stability while maintaining rendering quality and lip synchronization. The method uses editable AU trajectories as external controls rather than inferring emotion from audio alone. Code is available on \url{https://github.com/HMQH/Emozone-Talker}.

\end{abstract}

\ccsdesc[300]{Computing methodologies~Computer vision}

\keywords{3D Gaussian Splatting, Talking Head Synthesis, AU-Conditioned Expression Control, Facial Action Units}
\begin{teaserfigure}
  \includegraphics[width=\textwidth]{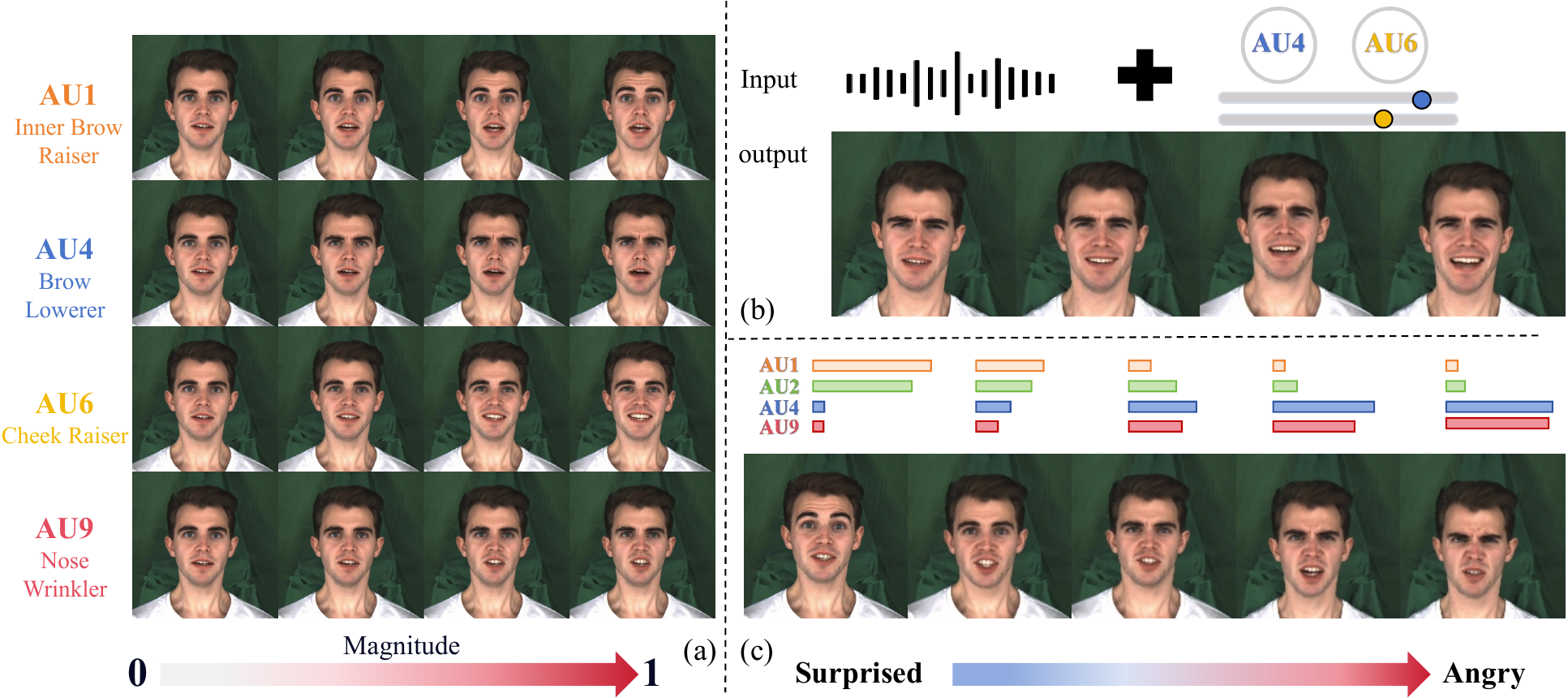}
  \caption{Fine-grained AU control with temporal coherence.
  	(a) Individual AU editing with controllable intensity.
  	(b) Coordinated synthesis under conflicting AU cues.
  	(c) Smooth transitions between affective states with reduced jitter.}
  \Description{The figure presents a structured comparison of facial expression control across multiple conditions. The left panel shows isolated manipulation of specific facial regions, where changes are localized to brows, cheeks, or nose without affecting other areas, indicating spatial disentanglement. Intensity increases progressively across columns, demonstrating continuous control rather than discrete switching. The upper-right panel combines audio input with multiple control signals, showing that simultaneous activation of different facial components results in coherent and non-conflicting expressions. The lower-right panel illustrates temporal progression across frames, where facial configurations evolve smoothly between distinct emotional states while preserving identity consistency and avoiding abrupt changes. Visual indicators highlight that different regions respond independently yet remain coordinated over time.}
  \label{fig:teaser}
\end{teaserfigure}


\maketitle

\section{Introduction}

Audio-driven talking head synthesis has attracted sustained attention
due to its broad applications in virtual avatars, human--computer
interaction, filmmaking, and digital content creation. Recent advances
in Neural Radiance Fields (NeRF), explicit dynamic geometry, and
3D Gaussian Splatting (3DGS) have substantially improved rendering
fidelity, geometric consistency, and inference efficiency
\cite{DBLP:conf/iccv/GuoCLLBZ21,
	DBLP:conf/cvpr/ZhangZLHLWGCL024,
	10.1145/3664647.3681627,
	DBLP:conf/eccv/LiZBZNZG24}.
However, most existing systems primarily focus on speech-driven
realism and lip synchronization, while providing limited support for
editable and region-specific facial expression control.

Existing approaches commonly enhance expressiveness through global
emotion labels, style embeddings, text prompts, reference signals, or
continuous affect representations
\cite{DBLP:conf/iccv/GanYYSY23,
	DBLP:conf/iccv/TanJP23,
	DBLP:conf/fgr/WangF25,
	DBLP:journals/corr/abs-2502-00654}.
Although these interfaces are effective for holistic emotional
modulation, they provide limited access to individual facial actions,
their spatial locations, and their activation intensities. In contrast,
Action Units (AUs), defined by the Facial Action Coding System (FACS)
\cite{friesen1978facial,DBLP:journals/pami/DonatoBHES99},
represent localized and semantically meaningful facial actions.
Continuous AU intensities therefore provide a more interpretable and
editable interface for fine-grained expression control.

Introducing explicit AU controls turns audio-driven facial animation into a multi-source control problem, in which speech and AU signals may exert competing influences over overlapping facial regions. Audio primarily drives the
lips, jaw, and surrounding articulation-related areas, whereas AUs may
also affect the cheeks, nasolabial folds, and mouth corners. Without
explicit regional responsibility assignment, implicit multimodal fusion
may suppress the requested AU response or disturb speech-driven
articulation. Second, frame-level AU trajectories extracted by facial
analysis tools often contain short-term fluctuations. Directly injecting
such trajectories into a deformation network may amplify the noise and
produce visible temporal jitter.

Recent methods do not fully address these issues. TalkingGaussian
\cite{DBLP:conf/eccv/LiZBZNZG24} improves structural persistence and
uses AUs as auxiliary training signals, but does not expose them as
explicit editable controls. EmoTalkingGaussian
\cite{DBLP:journals/corr/abs-2502-00654} introduces continuous
valence--arousal conditioning, but its control remains holistic rather
than region-specific. EmoDiffTalk
\cite{DBLP:journals/corr/abs-2512-05991} incorporates AU information
into an editable diffusion-based framework, yet still relies on implicit
conditioning without explicitly assigning modality preference across
overlapping facial regions. Therefore, achieving localized AU control
while maintaining stable temporal dynamics and accurate articulation
remains challenging.

To address these problems, we propose \emph{EmoZone-Talker}, an
AU-conditioned 3DGS framework for editable regional expression control.
The model accepts continuous AU trajectories as external control
signals, which can be extracted from reference videos, interpolated from
user-specified keyframes or sliders, or generated by an external
text/emotion-to-AU model. It does not infer expressions from audio
alone. Our Synergy Zones with Prioritized Attention Bias (SZ-PAB)
introduces soft facial-zone priors into cross-modal attention, biasing
audio toward articulation-related regions and AUs toward upper-face
regions while retaining a flexible synergy zone. We further introduce a
Channel-Independent Temporal AU Encoder (CIT-AE), which applies
channel-wise local temporal modeling before multimodal fusion to reduce
high-frequency AU fluctuations. The resulting audio and AU features
jointly condition spatially varying 3D Gaussian deformations.

The main contributions of this work are as follows:

\begin{itemize}
	\item We formulate AU-conditioned talking head synthesis as
	editable regional control under overlapping speech and expression
	influences, and identify spatial interference and noisy temporal AU
	conditioning as two key failure modes.
	
	\item We propose SZ-PAB, which introduces soft facial-zone priors
	into cross-modal attention to improve region-level AU controllability
	while retaining audio-driven articulation and cross-region
	flexibility.
	
	\item We introduce CIT-AE, a channel-wise local temporal encoder
	whose outputs are combined into a shared AU embedding, reducing
	AU-driven temporal jitter before 3D Gaussian deformation.
\end{itemize}

\section{Related Work}

\begin{figure*}[!t]
	\centering
	\includegraphics[width=\textwidth]{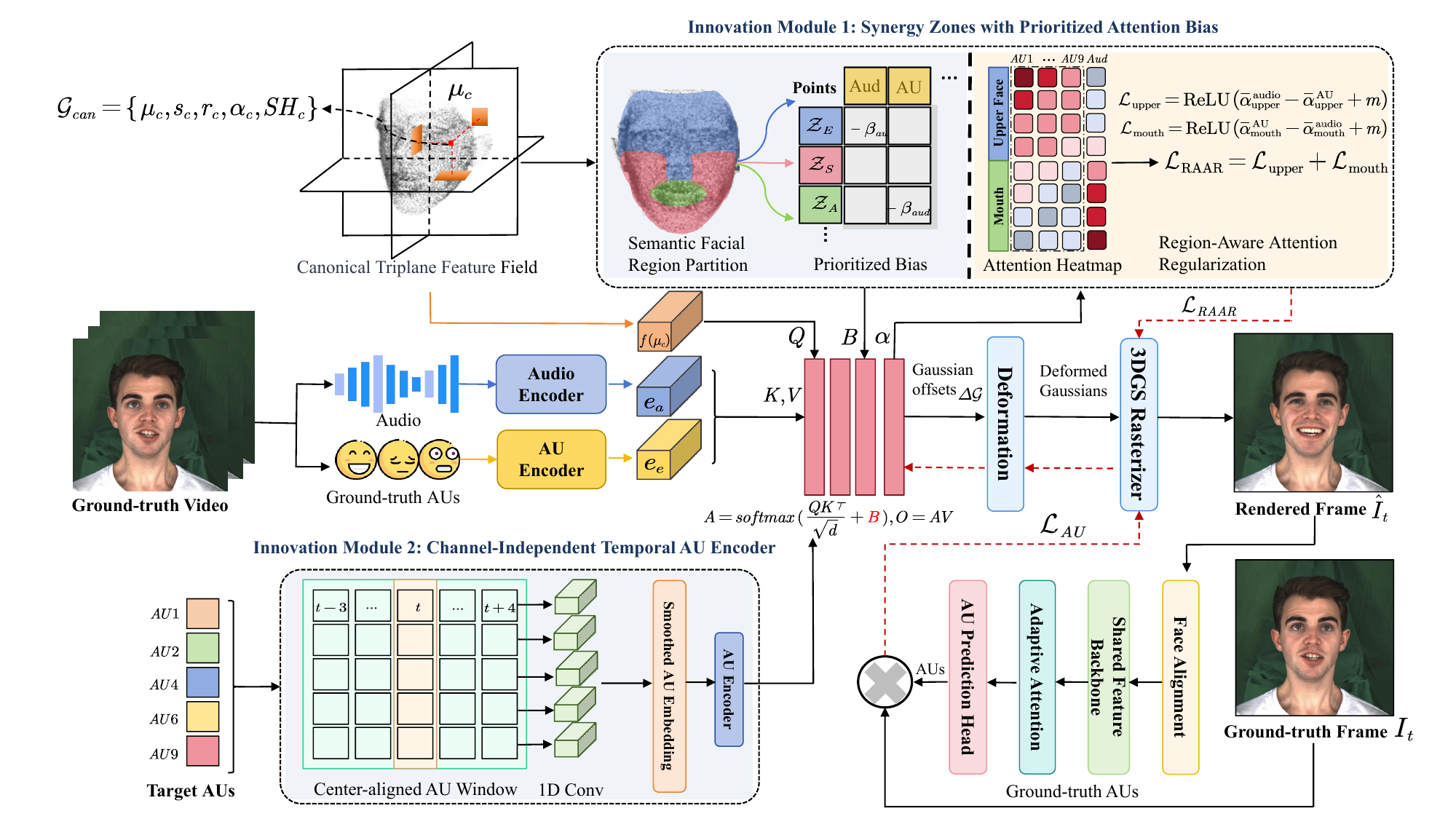}
	\caption{Overview of EmoZone-Talker. Audio, AU, and tri-plane features jointly predict Gaussian deformations. SZ-PAB applies region-aware attention bias, CIT-AE stabilizes AU dynamics, and AU-consistency supervision preserves expression fidelity.}
	\Description{The diagram depicts the internal data flow and interactions between components in a multi-branch system. Inputs include video frames, audio signals, and structured control parameters that are encoded into separate feature representations. These features are aligned and fused through attention-based mechanisms, where query, key, and value projections integrate modality-specific information. A region-aware attention process assigns different importance to facial areas, guided by a predefined semantic partition and bias maps that emphasize regions such as the upper face or the mouth. Deformation parameters are then predicted and applied to a set of primitive representations to generate rendered outputs. A parallel temporal modeling pathway processes sequences of control signals using sliding windows and channel-wise operations to capture local temporal dependencies and ensure smooth transitions. Additionally, an auxiliary consistency branch compares predicted control signals with reference targets, enforcing alignment through reconstruction and classification losses. The overall structure highlights the separation of spatial control, temporal modeling, and supervision signals, while maintaining coordinated information flow across modules.}
	\label{fig:fig2}
\end{figure*}

\subsection{Audio-Driven Talking Head Synthesis}
Audio-driven talking head synthesis has progressed from 2D pipelines to 3DMM-, NeRF-, and 3DGS-based methods. Early 2D approaches, such as MakeItTalk~\cite{10.1145/3414685.3417774} and SadTalker~\cite{DBLP:conf/cvpr/ZhangCWZSGSW23}, use explicit intermediate representations for controllability but lack rendering quality and 3D consistency. NeRF-based methods improve realism, view consistency, temporal stability, and lip synchronization by mapping audio to dynamic radiance fields~\cite{DBLP:conf/iccv/GuoCLLBZ21,DBLP:conf/iclr/YeJ0LHZ23,DBLP:journals/corr/abs-2305-00787,DBLP:journals/ijcv/TangWZCHHLLZW25,DBLP:conf/iccv/LiZ00023,DBLP:conf/cvpr/PengHS0ZZ00F24}. More recently, 3DGS enables high-fidelity real-time rendering. GaussianTalker~\cite{10.1145/3664647.3681627} predicts audio-driven Gaussian deformation, while TalkingGaussian~\cite{DBLP:conf/eccv/LiZBZNZG24} introduces smooth deformation and auxiliary AU supervision.

\subsection{Emotion-Conditioned Talking Head Generation}
Emotion- and style-conditioned systems use labels, embeddings, references, or diffusion guidance to enhance expressiveness~\cite{DBLP:conf/cvpr/JiZWWLCX21,10.1145/3528233.3530745,DBLP:conf/aaai/MaWHFL0D023,DBLP:conf/iccv/GanYYSY23,DBLP:conf/eccv/TianWZB24}. DICE-Talk~\cite{10.1145/3746027.3755286} models identity and emotion, while EmoTalk3D~\cite{DBLP:conf/eccv/HeJGLDHYZMXWZCZ24} and EmoTalkingGaussian~\cite{DBLP:journals/corr/abs-2502-00654} extend emotion control to 3D synthesis. These interfaces primarily describe holistic affect rather than editable local actions.

\subsection{Fine-Grained Expression Control via Action Units}

Action Units (AUs) decompose facial expressions into localized, semantically meaningful activations~\cite{friesen1978facial}. They complement higher-level interfaces such as text-guided style control~\cite{ma2024talkcliptalkingheadgeneration}, multimodal coarse-to-fine control~\cite{chen2025cafe}, and timeline-based motion control~\cite{ma2025exploringtimelinecontrolfacial} by exposing continuous intensity and region-specific editing. AU-based methods use landmarks, two-stage synthesis, or diffusion to control facial motion~\cite{DBLP:journals/corr/abs-2509-19749,DBLP:journals/corr/abs-2602-09534,DBLP:journals/corr/abs-2512-05991,xie2026finegrainedfacialcontrol3d}. Our focus is narrower: explicit regional responsibility and temporal stabilization for editable AU trajectories in a real-time, per-identity 3DGS deformation model.

\section{Method}
\subsection{Overview}

Our model maps audio, AU, camera, and canonical tri-plane features to frame-specific Gaussian offsets and renders the deformed Gaussians (Fig.~\ref{fig:fig2}). The editable control is a 5D continuous intensity vector comprising AU1 (Inner Brow Raiser), AU2 (Outer Brow Raiser), AU4 (Brow Lowerer), AU6 (Cheek Raiser), and AU9 (Nose Wrinkler), each on a 0--5 scale. These AUs cover representative upper- and mid-face actions; AU6 is also challenging because cheek motion can affect the mouth corners.

At inference time, AU trajectories can be extracted from reference videos, interpolated from sparse user keyframes or sliders, or supplied by an external text/emotion-to-AU front-end. They are editable controls, not expressions inferred by our model from audio alone. CIT-AE stabilizes the trajectories, SZ-PAB biases audio--AU fusion by region, and the auxiliary AU branch supervises whether rendered frames reflect the requested controls.

\paragraph{Gaussian deformation data flow.}
We represent the canonical subject as a set of 3D Gaussians~\cite{kerbl20233d,10.1145/3664647.3681627,DBLP:conf/eccv/LiZBZNZG24},
\begin{equation}
\mathcal{G}_{\mathrm{can}}=\{g_i\}_{i=1}^{N},\quad
g_i=(\mu_i,s_i,r_i,\alpha_i,SH_i),
\end{equation}
where position $\mu_i$, scale $s_i$, rotation $r_i$, opacity $\alpha_i$, and spherical-harmonic coefficients $SH_i$ describe geometry and appearance. For every Gaussian center, a multi-resolution tri-plane field supplies a spatial feature
\begin{equation}
\mathbf{f}_i=\Phi(\mu_i)\in\mathbb{R}^{d_f}.
\end{equation}
These features form the transformer queries. The temporal audio encoder, CIT-AE, and camera encoder produce $\mathbf{E}_{audio}$, $\mathbf{E}_{AU}$, and $\mathbf{E}_{cam}$, respectively, which form the condition tokens:
\begin{equation}
Q=\mathbf{F}_{spatial},\qquad
K,V=\operatorname{Concat}(\mathbf{E}_{audio},\mathbf{E}_{AU},\mathbf{E}_{cam}).
\end{equation}
SZ-PAB adds its region-dependent bias inside cross-attention. The fused feature of each Gaussian is decoded into frame-specific offsets
\begin{equation}
\Delta\mathcal{G}_t=\{\Delta\mu_t,\Delta r_t,\Delta s_t,
\Delta\alpha_t,\Delta SH_t\},
\end{equation}
which are applied to $\mathcal{G}_{\mathrm{can}}$ before differentiable rasterization. Thus the AU trajectory directly conditions Gaussian deformation rather than serving only as an auxiliary label. The five selected AUs cover complementary brow, cheek, and nose actions while avoiding direct duplication of the speech-driven jaw and lip controls; AU6 deliberately retains cheek--mouth coupling as a stress case for the soft regional prior.

\subsection{Synergy Zones with Prioritized Attention Bias (SZ-PAB)}

Audio-driven facial animation can be formulated as AU-conditioned face generation:
\begin{equation}
y = f(x_{audio}, x_{AU}),
\end{equation}
where speech drives articulation, and AUs explicitly control facial expressions.
A key challenge is cross-modal interference, as both signals affect overlapping facial regions, leading to ambiguous and entangled control. To address this, we introduce \textbf{SZ-PAB}, which imposes a region-wise inductive bias over cross-modal interactions. Specifically, we encourage a structured factorization of the generation process:
\begin{equation}
p(y | x_{audio}, x_{AU}) \approx \prod_{r} p(y_r | x_{audio}^{(r)}, x_{AU}^{(r)}),
\end{equation}
so that different facial regions preferentially respond to modality-specific signals. 
To instantiate this factorization, we introduce a region-wise
decomposition that encourages region-dependent modality preference
across facial regions. This factorization serves as a motivating approximation rather than
assuming strict independence among facial regions.

\paragraph{Synergy Zone Partitioning.}
Using landmarks detected on the canonical frame, we define an audio-dominant zone $\mathcal{Z}_A$ (mouth), an expression-dominant zone $\mathcal{Z}_E$ (upper face), and a synergy zone $\mathcal{Z}_S$ (transition regions). The boundaries remain fixed during training and inference:
\begin{equation}
L(\mu_i) \in \{\mathcal{Z}_A, \mathcal{Z}_E, \mathcal{Z}_S\}.
\end{equation}
This partition is a soft structural prior for modality preference rather than a strict anatomical decomposition.

\paragraph{Prioritized Attention Bias.}
We incorporate this prior into cross-attention via a learnable bias:
\begin{equation}
\text{Attn}(Q, K, V) = \text{softmax}\left(\frac{QK^\top}{\sqrt{d}} + B\right)V,
\end{equation}
where $B$ is a learnable region-dependent bias. It encourages modality preference without masking either modality: $\mathcal{Z}_A$ favors audio tokens, $\mathcal{Z}_E$ favors AU tokens, and $\mathcal{Z}_S$ remains a flexible transition region. Learned attention can therefore preserve cross-zone facial coupling while reducing interference.

\paragraph{Region-Aware Regularization.}
\label{subsec:RAAR}
To further stabilize region-wise attention allocation, we introduce \textbf{Region-Aware Attention Regularization (RAAR)}, which enforces modality preference at the region level. Let $\alpha_{i,j}$ denote the attention weight from Gaussian $i$ to token $j$. We compute the average attention response of each region to different modalities. For example, the upper-face response to audio tokens is defined as:
\begin{equation}
\bar{\alpha}^{audio}_{upper} = \frac{1}{|\mathcal{Z}_E|} \sum_{i \in \mathcal{Z}_E} \sum_{j \in \text{audio}} \alpha_{i,j}
\end{equation}

Based on these statistics, we impose a margin-based constraint:
\begin{equation}
\mathcal{L}_{RAAR} = \mathcal{L}_{upper} + \mathcal{L}_{mouth},
\end{equation}
where $\mathcal{L}_{upper}$ encourages AU attention to dominate over audio in the upper-face region, while $\mathcal{L}_{mouth}$ enforces the opposite preference in the mouth region using hinge-style losses with margin $m$.
This design reinforces region-consistent modality preference while maintaining flexibility during training.

\subsection{Channel-Independent Temporal AU Encoder (CIT-AE)}

Frame-level AU signals can be modeled as noisy temporal observations:
\begin{equation}
\mathbf{a}_t = \mathbf{s}_t + \boldsymbol{\epsilon}_t,
\end{equation}
where $\mathbf{s}_t$ denotes the underlying expression dynamics and $\boldsymbol{\epsilon}_t$ represents high-frequency measurement noise. Direct conditioning on $\mathbf{a}_t$ can introduce jitter. CIT-AE therefore applies learnable local temporal filtering before multimodal fusion. For each frame $t$, it constructs a center-aligned temporal window:
\begin{equation}
\mathbf{A}_t = [\mathbf{a}_{t-h}, \dots, \mathbf{a}_t, \dots, \mathbf{a}_{t+h-1}] \in \mathbb{R}^{T \times K},
\end{equation}
Rather than mixing channels within the temporal operator, CIT-AE processes the sequence of each AU $k$ with a dedicated one-dimensional convolution:
\begin{equation}
\mathbf{h}^{(k)}_t=\operatorname{Conv1D}^{(k)}(\mathbf{A}^{(k)}_t),
\end{equation}
where $\mathbf{A}^{(k)}_t\in\mathbb{R}^{T}$ is the local trajectory for channel $k$. The response aligned to the center frame is retained,
\begin{equation}
\tilde{a}^{(k)}_t=\mathbf{h}^{(k)}_t[\mathrm{center}],
\end{equation}
and the channel responses are then concatenated and projected by a shared AU encoder:
\begin{equation}
\tilde{\mathbf{a}}_t=
[\tilde{a}^{(1)}_t,\ldots,\tilde{a}^{(K)}_t]\in\mathbb{R}^{K},
\qquad \mathbf{E}_{AU}=\operatorname{Enc}_{AU}(\tilde{\mathbf{a}}_t).
\end{equation}

This ordering separates per-channel temporal filtering from subsequent cross-channel coordination: each AU first retains its own local dynamics, and only the resulting center responses enter the shared representation used by the deformation transformer. CIT-AE is therefore not an audio--AU temporal disentanglement module; its scoped role is to reduce noisy AU fluctuations before joint deformation prediction.

\begin{figure*}[!t]
	\centering
	\includegraphics[width=\textwidth]{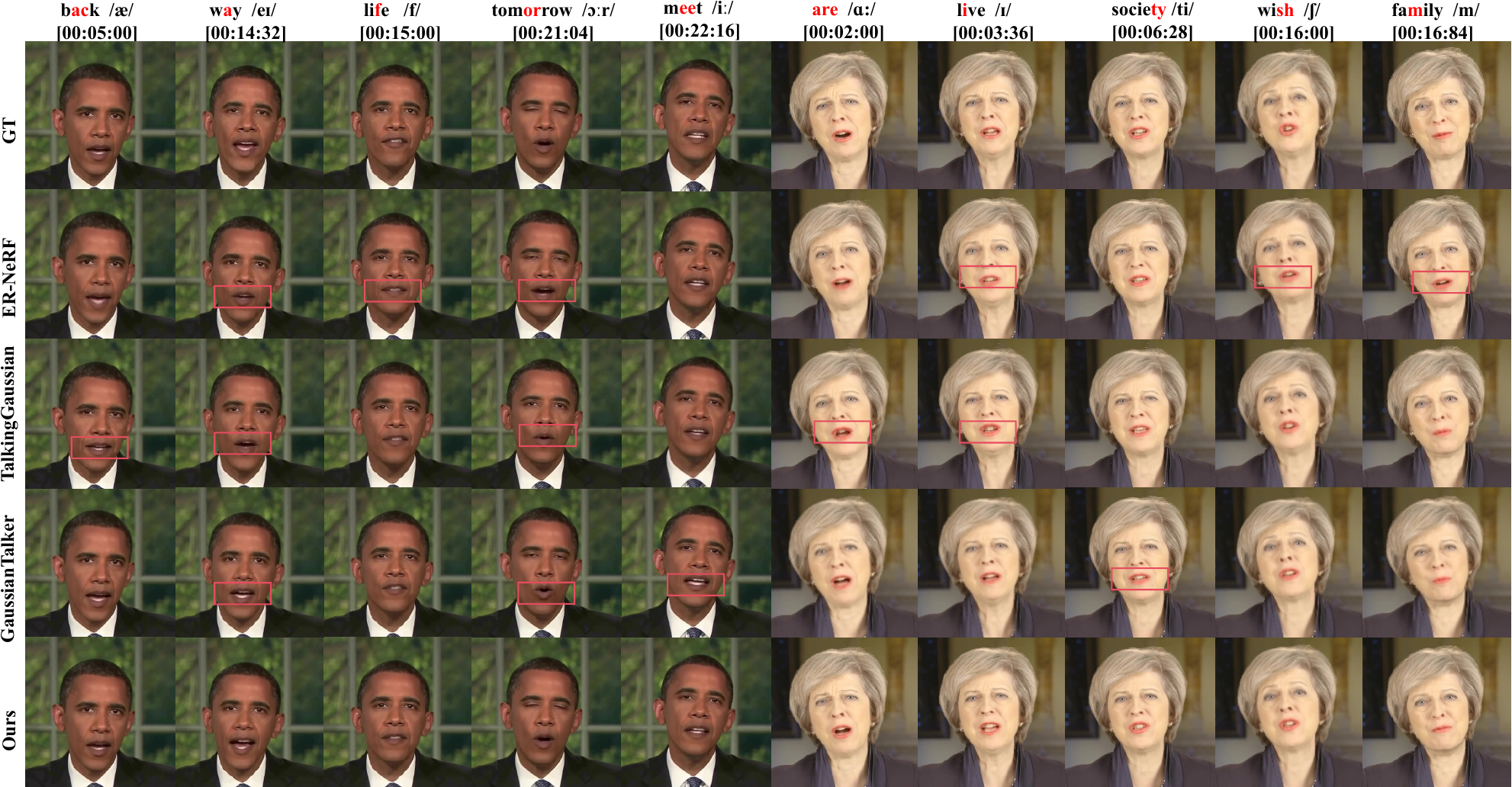}
	\caption{Qualitative comparison in the self-reconstruction scenario, where each subject is driven by their own speech. The regions highlighted by red solid boxes indicate mismatched lip articulation. }
	\Description{ The columns show selected frames aligned to speech content, with each frame labeled by a timestamp and the corresponding phoneme or word segment to indicate the intended audio-visual synchronization. Each row presents self-reconstruction results from a different method under the same speaking condition, enabling direct visual comparison across models. The red boxes highlight the mouth region and emphasize articulation errors at critical phonetic moments, especially when an open-vowel sound should produce a clearly open mouth. In some competing methods, the lip motion is inconsistent with the audio, such as insufficient mouth opening or a nearly closed mouth during vowel production, whereas the proposed method maintains more accurate and temporally coherent articulation. The two subjects on the left and right further illustrate that this behavior is evaluated across different identities, helping show robustness to speaker variation. The frame-to-frame progression also reveals differences in motion smoothness and continuity, which are important for assessing realistic talking-face generation.}
	\label{fig:fig3}
\end{figure*}

\subsection{Joint AU Detection and Face Alignment (JAA)-Based AU Supervision}
\label{JAA-Based AU Supervision}

Introducing AU conditions does not guarantee that the rendered motion faithfully reflects the intended expression semantics, as the deformation network may underutilize AU signals or entangle them with dominant audio features. To address this, we introduce a JAA-based supervision branch that explicitly enforces AU consistency during training.

For each frame, we obtain target AU labels $\mathbf{a}^{gt}_t$ using a pre-trained AU detector~\cite{DBLP:conf/fgr/BaltrusaitisZLM18}. Given the rendered frame $\hat{I}_t$, we estimate its AU responses using a pre-trained JAA network~\cite{shao2018deep_jaa}, which serves as an external AU estimator for measuring expression consistency:
\begin{equation}
	\hat{\mathbf{a}}_t = \Psi_{JAA}(\hat{I}_t).
\end{equation}

We then define an AU consistency loss:
\begin{equation}
  \mathcal{L}_{\text{AU}} = \|\hat{\mathbf{a}}_t - \mathbf{a}^{\text{gt}}_t\|_1.
\end{equation}

This supervision encourages the model to produce expression-consistent facial responses while maintaining photorealism. Notably, the JAA branch is only used during training and introduces no additional inference cost. It can be viewed as an expression-aware perceptual constraint that provides semantic supervision beyond pixel-level reconstruction.

\begin{figure*}[!t]
	\centering
	\includegraphics[width=\textwidth]{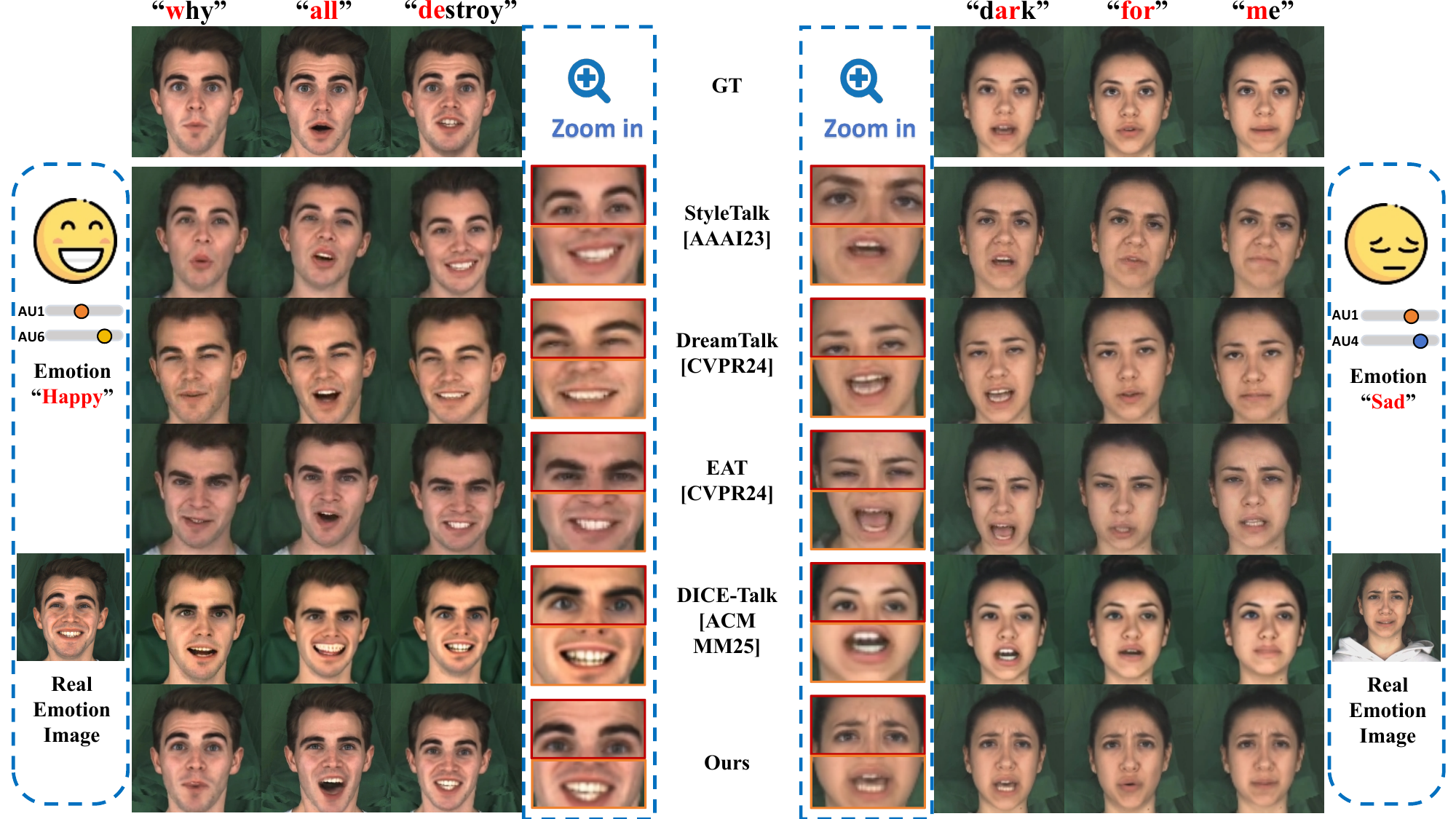}
	\caption{Qualitative comparison under the same audio and target
		emotion, using each method's native control interface.}
	\Description{The figure organizes results by emotion category, with the left panel corresponding to a “happy” target expression and the right panel to a “sad” target expression, each driven by the same input speech content indicated above the frames. For each subject, multiple methods are compared row-wise, including several baselines and the proposed approach, while the ground truth sequence is provided for reference. Zoomed-in regions focus on the eye and mouth areas, highlighting subtle expression details such as eye narrowing, eyebrow movement, and lip curvature that are critical for conveying affect. The annotated action units (e.g., AU1, AU4, AU6) illustrate the facial muscle activations associated with each emotion, providing an interpretable link between visual appearance and expression semantics. It can be observed that some methods fail to consistently reproduce the intended emotional cues, either exaggerating or weakening key facial movements, whereas the proposed method better captures both global expression consistency and fine-grained local details. Additionally, the comparison with real emotion images at the sides offers a visual reference for natural expression intensity and distribution, enabling a more comprehensive assessment of realism and emotional fidelity.}
	\label{fig:fig4}
\end{figure*}

\subsection{Training Objectives}
\label{sec:training_objectives}
The overall training objective combines reconstruction fidelity, lip synchronization, AU consistency, and region-aware regularization:
\begin{equation}
	\mathcal{L}_{\text{total}} = \mathcal{L}_{\text{rec}} + \lambda_{\text{sync}} \mathcal{L}_{\text{sync}} + \lambda_{\text{AU}} \mathcal{L}_{\text{AU}} + \lambda_{\text{RAAR}} \mathcal{L}_{\text{RAAR}}.
\end{equation}

Here, $\mathcal{L}_{\text{rec}}$ enforces photometric and perceptual consistency, $\mathcal{L}_{\text{sync}}$ ensures speech-lip alignment using SyncNet~\cite{DBLP:conf/accv/ChungZ16a}, $\mathcal{L}_{\text{AU}}$ enforces expression semantics via JAA-based supervision, and $\mathcal{L}_{\text{RAAR}}$ regularizes region-wise attention allocation. We empirically set $\lambda_{\text{sync}} = 0.1$, $\lambda_{\text{AU}} = 0.05$, and $\lambda_{\text{RAAR}} = 0.01$ in all experiments.

Together, these losses optimize rendering fidelity, speech--lip alignment, regional expression control, and temporal coherence.




\section{Experiments}
\label{sec:experiments}

\subsection{Experimental Setting}
\label{subsec:setup}

\paragraph{Datasets.}
We evaluate self-reconstruction on the neutral subset containing Obama, May, Lieu, and Macron, following TalkingGaussian~\cite{DBLP:conf/eccv/LiZBZNZG24}. We further evaluate AU-conditioned generation on MEAD~\cite{DBLP:conf/eccv/WangWSYWQHQL20}, which contains 60 speakers performing eight emotion categories at three intensity levels. Emotion labels are used only to organize this evaluation; the proposed model is driven by the corresponding continuous AU trajectories. Videos use a 10:1 training/test split~\cite{10.1145/3664647.3681627,DBLP:conf/eccv/LiZBZNZG24}, $256\times256$ resolution, 25 FPS, and 16 kHz audio. JAANet supplies training supervision, whereas OpenFace~\cite{DBLP:conf/fgr/BaltrusaitisZLM18} is used for evaluation to reduce evaluator--optimizer coupling. The RAVDESS examples test qualitative transfer only and are not included in the quantitative claims.

\paragraph{Implementation Details.}
Our model is implemented in PyTorch and optimized using Adam~\cite{DBLP:journals/corr/KingmaB14}. Training consists of a coarse stage followed by a fine stage for lip and depth refinement. The learning rate decays exponentially from $1\times10^{-4}$ to $1\times10^{-5}$. 3D Gaussian densification~\cite{10.1145/3664647.3681627} is performed every 100 iterations between 1k and 7k steps.

\paragraph{Evaluation Metrics.}
We report PSNR and SSIM~\cite{DBLP:journals/tip/WangBSS04,DBLP:conf/icpr/HoreZ10} for reconstruction, LPIPS~\cite{DBLP:conf/cvpr/ZhangIESW18} for perceptual distance, landmark distance (LMD)~\cite{DBLP:conf/cvpr/ChenMDX19} for facial geometry, and identity similarity (CSIM)~\cite{DBLP:conf/cvpr/SchroffKP15}. Lip synchronization is measured by both SyncNet confidence and lip-sync error distance (Sync/LSE-D)~\cite{DBLP:conf/accv/ChungZ16a}; higher Sync and lower LSE-D are preferred. AUE-U/L are mean absolute errors between target and detected AU intensities in the upper and lower facial groups. AU-Jerk is the mean magnitude of the second-order temporal difference of the detected AU trajectory, so it measures frame-to-frame stability rather than expression correctness by itself. E-Score uses a pretrained AffectNet classifier~\cite{DBLP:journals/taffco/MollahosseiniHM19}. Because AU supervision and AU metrics both rely on detectors, AUE and AU-Jerk remain proxy measurements of detector agreement. We therefore interpret them jointly with rendering, synchronization, qualitative, and user-study evidence.


\paragraph{Comparisons.}
Self-reconstruction baselines are ER-NeRF~\cite{DBLP:conf/iccv/LiZ00023}, GaussianTalker~\cite{10.1145/3664647.3681627}, and TalkingGaussian~\cite{DBLP:conf/eccv/LiZBZNZG24}. MEAD baselines are StyleTalk~\cite{DBLP:conf/aaai/MaWHFL0D023}, EAT~\cite{DBLP:conf/iccv/GanYYSY23}, DreamTalk~\cite{DBLP:journals/corr/abs-2312-09767}, and DICE-Talk~\cite{10.1145/3746027.3755286}. They receive coarse emotion controls, whereas ours receives AU trajectories. The cross-interface comparison tests practical controllability;
Table~\ref{tab:ablation_arch} isolates architectural gains under matched AU controls.

\paragraph{Evaluation Protocol.}
The three evaluation settings answer complementary questions.Self-reconstruction evaluates whether the additional AU pathway preserves rendering fidelity, facial geometry, identity, and speech--lip synchronization. The MEAD control comparison evaluates the practical capability of complete systems under their native control
interfaces, but does not by itself isolate architectural improvements. We therefore use matched-input ablations, in which all variants receive the same AU trajectories, to separately examine the effects of SZ-PAB, CIT-AE, and continuous AU intensities. This separation prevents gains from a stronger control interface from being directly attributed to a specific architectural component.

\subsection{Quantitative Results}
\label{subsec:quantitative}

\paragraph{Self-Reconstruction.}
Table~\ref{tab:comparison} shows that our method has the lowest AUE-U on both datasets while retaining competitive Sync and the lowest MEAD LSE-D. On the neutral subset, AUE-U decreases from the second-best 0.206 to 0.156; on MEAD it decreases from 0.163 to 0.145. The method also obtains the lowest LMD on both sets, while LPIPS is not consistently best. This mixed pattern supports improved AU realization and geometry without claiming uniform superiority in every aspect of appearance. The MEAD LSE-D of 6.93, together with Sync 5.228, indicates that the added AU pathway does not remove usable lip synchronization. Runtime is 110.5 FPS, compared with 116.5 FPS for GaussianTalker, 105 FPS for TalkingGaussian, and 34.7 FPS for ER-NeRF.

\paragraph{MEAD Control.}
Table~\ref{tab:emotion_gen} reports the MEAD emotion-conditioned setting. Our AU-controlled model obtains an E-Score of 0.653 and CSIM of 0.841, compared with 0.552 and 0.651 for the strongest listed baseline values in the corresponding columns. The neutral/non-neutral split shows that the difference is not confined to only one subset. However, the baselines receive category-level emotion controls while ours receives continuous AU trajectories. The table therefore demonstrates the practical capability of the complete control interface in this setting; it does not isolate the contribution of SZ-PAB or CIT-AE. Those architectural effects are assessed in
Table~\ref{tab:ablation_arch} using matched AU inputs.

\subsection{Qualitative Results}
\label{subsec:qualitative}

\begin{table*}[t]
	\caption{Self-reconstruction results. Best and second-best values are bold and underlined.}
	\label{tab:comparison}
	\centering
    \setlength{\tabcolsep}{2.2pt}
	\small
	\begin{tabular}{l *{15}{c}}
		\toprule
        \multirow{2}{*}{\centering\arraybackslash\textbf{Method}}
		& \multicolumn{7}{c}{\textbf{Neutral Dataset}}
		& \multicolumn{8}{c}{\textbf{MEAD}} \\
		\cmidrule(lr){2-8}\cmidrule(lr){9-16}
		&PSNR$\uparrow$ &SSIM$\uparrow$ &LPIPS$\downarrow$ &LMD$\downarrow$ &Sync$\uparrow$ &CSIM$\uparrow$ &AUE-U$\downarrow$/L$\downarrow$ 
		&PSNR$\uparrow$ &SSIM$\uparrow$ &LPIPS$\downarrow$ &LMD$\downarrow$ &Sync$\uparrow$ &LSE-D$\downarrow$ &CSIM$\uparrow$ &AUE-U$\downarrow$/L$\downarrow$ \\
		\midrule
		ER-NeRF
		& 33.059 & 0.935 & \underline{0.027} & 2.269 & 5.554 & 0.930 & 0.271/0.228
		& 31.156 & 0.931 & \underline{0.033} & 2.649 & 4.492 & 8.07 & 0.864 & 0.355/0.376 \\
		GaussianTalker
		& 33.023 & 0.939 & 0.033 & 2.306 & 5.741 & \underline{0.946} & 0.256/0.228
		& 31.952 & \underline{0.946} & 0.047 & 2.410 & \underline{4.823} & \underline{7.21} & 0.913 & 0.338/0.330 \\
		TalkingGaussian
		& \underline{33.637} &\underline{0.940}  & \textbf{0.026} & \underline{2.013} & \textbf{5.919} & 0.945 & \underline{0.206/0.223}
		& \underline{32.769} & 0.942 & \textbf{0.029} & \underline{2.074} & 4.794 & 7.54 & \underline{0.914} & \underline{0.163/0.314} \\
		Ours
		& \textbf{34.732} & \textbf{0.952} & 0.030 & \textbf{1.951} & \underline{5.849} & \textbf{0.956} & \textbf{0.156/0.199}
		& \textbf{33.940} & \textbf{0.957} & 0.038 & \textbf{1.852} & \textbf{5.228} & \textbf{6.93} & \textbf{0.944} & \textbf{0.145/0.258} \\
		\bottomrule
	\end{tabular}
\end{table*}

\begin{table*}[!t]
	\caption{Ablation study of architecture components under matched AU inputs.}
	\label{tab:ablation_arch}
	\centering
	\setlength{\tabcolsep}{4pt}
	\begin{tabular}{lcccccccccc}
		\toprule
		Config
		& \textbf{AU Cond}
		& \textbf{SZ-PAB}
		& \textbf{CIT-AE}
		& PSNR $\uparrow$
		& LPIPS $\downarrow$
		& Sync $\uparrow$
		& AUE-U $\downarrow$
		& AUE-L $\downarrow$
		& AU-Jerk $\downarrow$ \\
		\midrule
		(A)  & -       & -       & -       & 32.879 & 0.042 & 5.741 & 0.309 & 0.314 & 0.137 \\
		(B)  & $\surd$ & -       & -       & 31.795 & 0.046 & 5.378 & 0.324 & 0.358 & 0.153 \\
		(C)  & $\surd$ & $\surd$ & -       & 33.947 & 0.037 & 5.569 & 0.152 & 0.235 & 0.126 \\
		(D)  & $\surd$ & -       & $\surd$ & 33.921 & 0.039 & 5.571 & 0.305 & 0.276 & 0.098 \\
		Ours & $\surd$ & $\surd$ & $\surd$
		& \textbf{33.960}
		& \textbf{0.034}
		& \textbf{5.712}
		& \textbf{0.143}
		& \textbf{0.231}
		& \textbf{0.095} \\
		\bottomrule
	\end{tabular}
\end{table*}

\begin{table}[!t]
	\caption{Ablation study of AU representations under matched AU inputs.}
	\label{tab:ablation_au}
	\centering
	\setlength{\tabcolsep}{4pt}
	\small
	\begin{tabular}{lccc}
		\toprule
		Representation
		& AUE-U $\downarrow$
		& AU-Jerk $\downarrow$
		& Sync $\uparrow$ \\
		\midrule
		Binary activation
		& 0.281
		& 0.101
		& 5.654 \\
		Continuous intensity
		& \textbf{0.143}
		& \textbf{0.095}
		& \textbf{5.712} \\
		\bottomrule
	\end{tabular}
\end{table}

\paragraph{Self-Reconstruction.}
Figure~\ref{fig:fig3} highlights selected articulation mismatches. In the shown frames, ours more closely follows the reference lip shape.

\paragraph{Emotion Control.}
Figure~\ref{fig:fig4} compares complete systems under the same audio and target emotion but different control interfaces. The AU-driven examples show clearer brow, cheek, and nose changes while retaining articulation. This qualitative evidence is subject to the interface difference in Sec.~\ref{subsec:setup}.

\subsection{Ablation Study}
\label{sec:ablation}

\begin{figure}[t]
	\centering
	\includegraphics[width=\columnwidth]{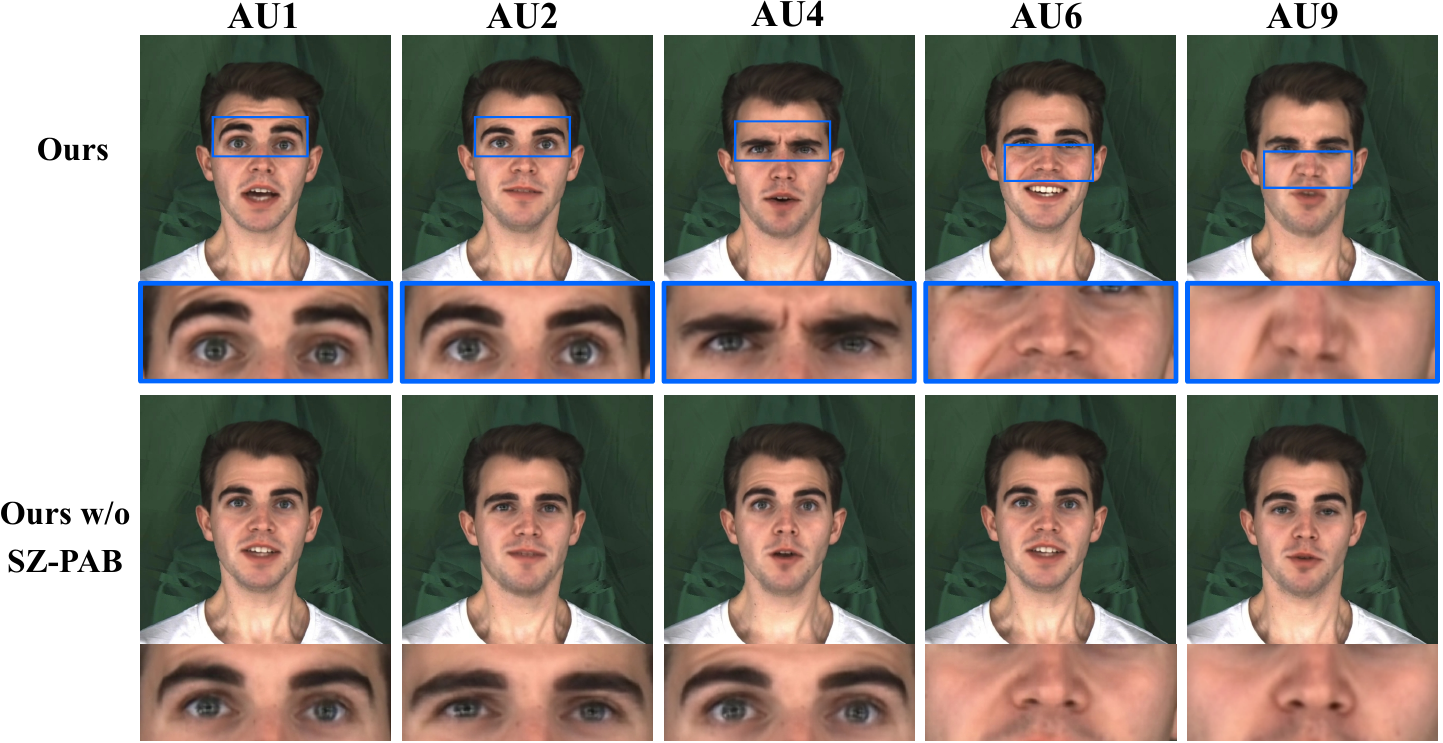}
	\caption{SZ-PAB ablation across AU1, AU2, AU4, AU6, and AU9.
		Blue boxes highlight the corresponding regional responses.}
	\Description{Regional AU responses with and without SZ-PAB.}
	\label{fig:fig5}
\end{figure}

\begin{figure}[t]
	\centering
	\includegraphics[width=\columnwidth, trim=0 9 0 7, clip]{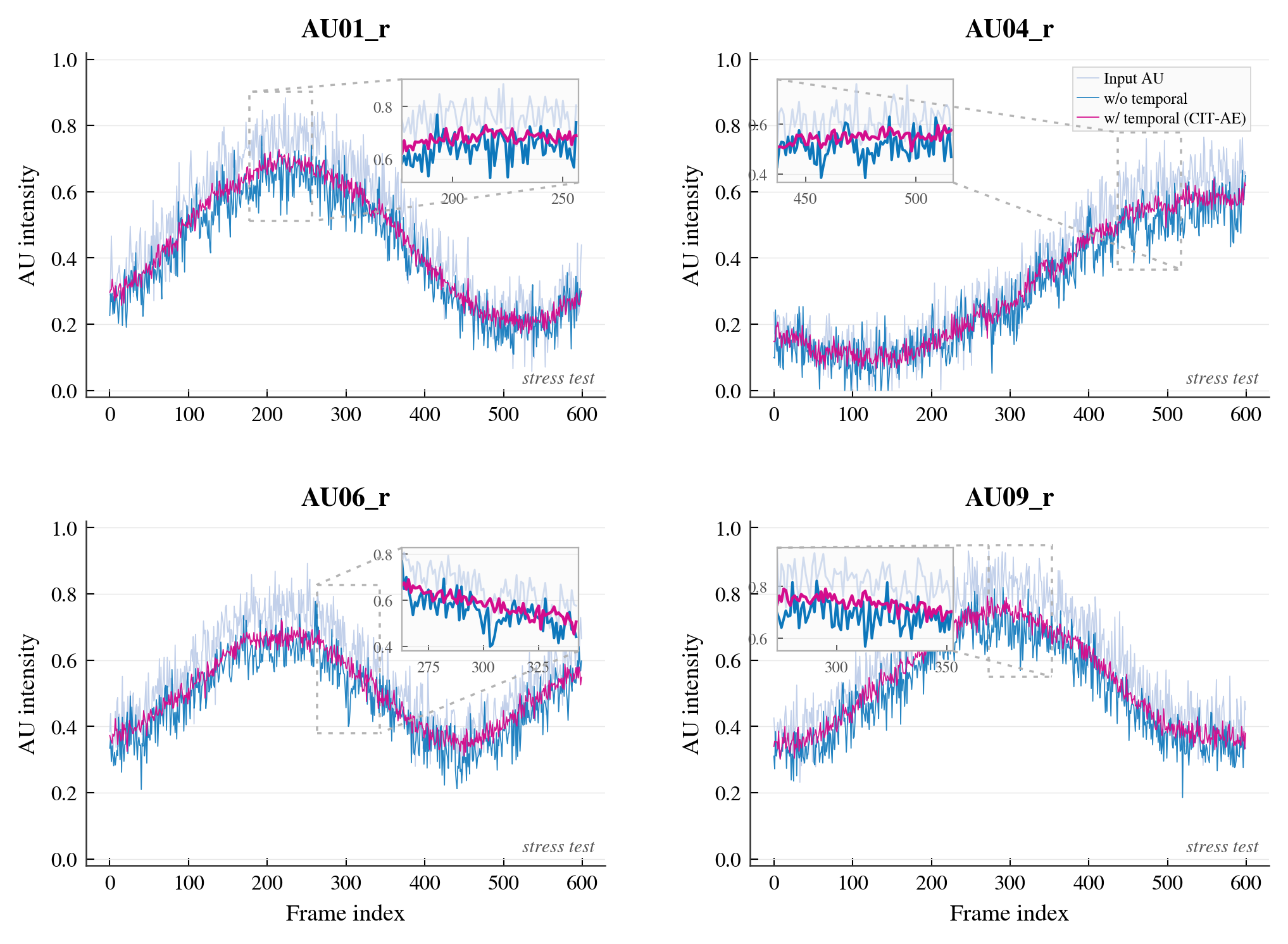}
	\caption{Temporal Stability Analysis under Stress Testing.}
	\Description{AU trajectories with and without CIT-AE under noisy inputs.}
	\label{fig:fig6}
\end{figure}

\begin{figure}[t]
    \centering
    \includegraphics[width=\columnwidth]{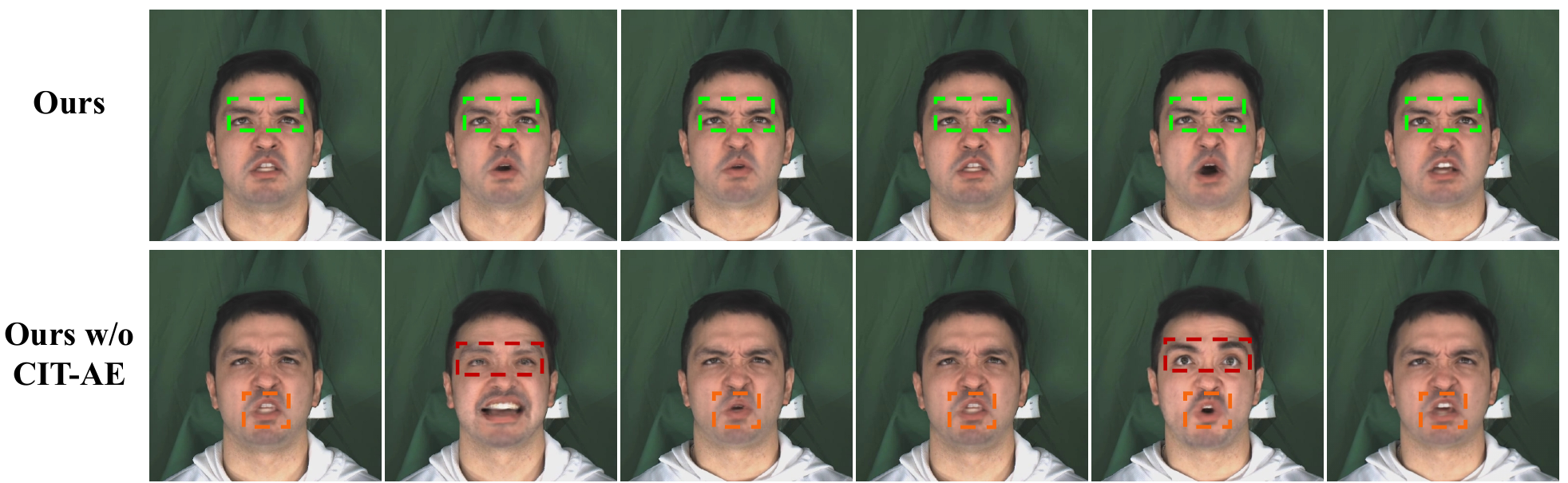}
    \caption{CIT-AE ablation. Without CIT-AE, activations become erratic (red) and temporally blurred (orange).}
    \Description{The top row shows the full model maintaining stable and consistent facial expressions across frames. The bottom row shows the variant without CIT-AE, with erratic upper-face activations and temporal blur in the mouth region under noisy AU inputs.}
    \label{fig:fig7}
\end{figure}

\begin{table}[htbp]
	\centering
	\caption{Quantitative comparison on emotion-conditioned generation on MEAD.}
	\label{tab:emotion_gen}
	\begin{tabular}{lcccccc}
		\toprule
		Method& E-Score $\uparrow$ & Neutral $\uparrow$& Non-neutral $\uparrow$ & CSIM $\uparrow$\\
		\midrule
        StyleTalk& 0.383 & 0.392 & 0.378 & 0.632 \\
        EAT& 0.552 & 0.368 & 0.613  & 0.616 \\
        DreamTalk& 0.526 & 0.585 & 0.501 & 0.606 \\
        DICE-Talk& 0.311 & 0.329 & 0.305 & 0.651 \\
		Ours& \textbf{0.653} & \textbf{0.601} & \textbf{0.685} &  \textbf{0.841} \\
		\bottomrule
	\end{tabular}
\end{table}


Tables~\ref{tab:ablation_arch} and~\ref{tab:ablation_au} report matched-input ablations of the architecture and AU representation, respectively; Figures~\ref{fig:fig5}--\ref{fig:fig7} provide complementary qualitative and temporal analyses.

\paragraph{Effect of regional attention bias.}
Comparing configurations (B) and (C), SZ-PAB reduces AUE-U from
0.324 to 0.152 under identical AU trajectories, corresponding to a
relative reduction of approximately 53\%. PSNR also increases from
31.795 to 33.947. Figure~\ref{fig:fig5} shows the corresponding
qualitative pattern: without SZ-PAB, brow and cheek responses are
visibly weaker in the selected examples, whereas the soft regional
prior makes the requested upper-face changes easier to observe.
These results support improved regional responsibility allocation
under matched inputs, rather than anatomically independent motion
or a hard separation between speech and expression.

\paragraph{Effect of temporal modeling.}
Comparing configuration (C) with the full model, CIT-AE reduces
AU-Jerk from 0.126 to 0.095, while AUE-U changes from 0.152 to
0.143. Under injected AU noise, the jitter reduction is 61\% with
CIT-AE versus 19\% without it (Fig.~\ref{fig:fig6}); the resulting
trajectories preserve the broad intensity trend while suppressing
more high-frequency variation. Figure~\ref{fig:fig7} provides a
complementary frame-level view of the resulting artifacts. These
results support learnable local AU filtering before multimodal
deformation, rather than perfect temporal control under every
transition.

\paragraph{Complementary effects of components.}
As shown in Table~\ref{tab:ablation_au}, continuous AU intensities outperform binary activations under the same architecture: AUE-U decreases from 0.281 to 0.143 and AU-Jerk from 0.101 to 0.095, while Sync increases from 5.654 to 5.712. This supports retaining the 0--5 intensity interface rather than thresholding each AU into a binary state. In the architecture ablation shown in Table~\ref{tab:ablation_arch}, adding raw AU conditioning alone degrades PSNR, Sync, and both regional AUE measures, indicating that naïve AU integration may introduce cross-modal interference. SZ-PAB primarily improves regional AU accuracy, whereas CIT-AE primarily reduces temporal jitter. Their combination achieves the lowest AUE-U, AUE-L, and AU-Jerk, together with the best rendering and synchronization results in the controlled architecture ablation. These results indicate that SZ-PAB and CIT-AE address complementary spatial and temporal failure modes and are most effective when used jointly.


\subsection{User Study}
\label{subsec:user_study}

\begin{table}[H]
	\centering
	\caption{Mean user-study ratings on five dimensions.}
	\label{tab:user_study}
	\resizebox{\linewidth}{!}{%
		\begin{tabular}{lccccc}
			\toprule
			Method & Quality & Emotion & Lip Sync & Identity & Naturalness \\
			\midrule
			StyleTalk & 3.38 & 2.68 & 3.65 & 2.34 & 2.98 \\
			EAT & 3.62 & 3.24 & 3.62 & 3.53 & 3.19 \\
			DreamTalk & 3.45 & 3.19 & 3.65 & 3.65 & 3.47 \\
			DICE-Talk & 3.12 & 2.82 & 3.45 & 2.47 & 3.21 \\
			Ours & \textbf{4.33} & \textbf{3.89} & \textbf{3.67} & \textbf{4.67} & \textbf{4.28} \\
			\bottomrule
		\end{tabular}%
	}
\end{table}

To complement our quantitative evaluation, we conducted a user study to assess the perceptual quality of emotion-conditioned generation. We recruited 20 participants (10 males, 10 females) who were not involved in this research. For each test case, participants viewed videos generated by our method and four competing emotional talking head methods (StyleTalk~\cite{DBLP:conf/aaai/MaWHFL0D023}, EAT~\cite{DBLP:conf/iccv/GanYYSY23}, DreamTalk~\cite{DBLP:journals/corr/abs-2312-09767}, DICE-Talk~\cite{10.1145/3746027.3755286}) in randomized order. They were asked to rate each video on a 5-point Likert scale (1 = poor, 5 = excellent) across five dimensions: overall quality, emotion accuracy, lip synchronization, identity preservation, and expression naturalness.

\paragraph{Results.}
Table~\ref{tab:user_study} reports the highest mean score for ours
in each rated dimension. The largest margins appear in identity
preservation and expression naturalness, while the lip-sync means
are close (3.67 for ours versus 3.65 for StyleTalk and DreamTalk).
This pattern is consistent with the quantitative evidence that AU
control can improve expression realization without producing a
large lip-sync advantage. We report means because per-participant
responses and confidence intervals are not available in the current
study; the results should not be interpreted as a significance test.

\section{Conclusion}
\label{sec:conclusion}

We presented \textbf{EmoZone-Talker}, an AU-conditioned framework for regional expression control in audio-driven 3DGS talking heads. The model accepts editable continuous AU trajectories and maps them, together with audio and camera conditions, to spatially varying Gaussian deformations. SZ-PAB introduces soft spatial responsibility priors, while CIT-AE applies channel-wise local filtering before shared multimodal deformation. Controlled ablations, temporal stress tests, and complementary perceptual,
rendering, and synchronization evaluations show improved AU realization
and temporal stability while retaining competitive visual quality and
lip synchronization.

\paragraph{Limitation}
The fixed canonical zones are approximate soft priors. Strong AU6 activation can affect the mouth corners, and AU1/AU2 share forehead musculature, so facial actions are neither anatomically independent nor perfectly confined to one zone. The learned attention may compensate for some coupling, but the current partition cannot adapt its boundary to subject-specific geometry. AUE and AU-Jerk are detector-based proxies and can inherit errors from OpenFace or JAANet; the qualitative RAVDESS examples likewise do not establish broad cross-dataset robustness. The renderer does not explicitly model teeth or tongue, which can limit oral detail under large mouth openings. Future work should investigate adaptive zones, uncertainty-aware AU supervision, broader perceptual evaluation, and explicit oral geometry.

\begin{acks}
This work is supported by the Shandong Natural Science Foundation Outstanding Young Scientist Fund (Grant No.~ZR2024YQ071) and the Independent Innovation Research Program for Science and Engineering (Grant No.~25CX06081A).
\end{acks}

\bibliographystyle{ACM-Reference-Format}
\balance
\bibliography{acmart}
\end{document}